\documentclass{article} % For LaTeX2e
\usepackage{iclr2025_conference,times}

\usepackage{amsmath,amsfonts,bm}

\def\eqref#1{equation~\ref{#1}}
\def\1{\bm{1}}

\DeclareMathAlphabet{\mathsfit}{\encodingdefault}{\sfdefault}{m}{sl}
\SetMathAlphabet{\mathsfit}{bold}{\encodingdefault}{\sfdefault}{bx}{n}

\usepackage{booktabs,hyperref}
\usepackage{url}
\iclrfinalcopy

\title{Two for One: Privileged
Learning for Resource-Constrained Screening}

\author{Lisa Y.W. Tang\\
University of British Columbia - Faculty of Medicine}

\begin{document}

\maketitle

\begin{abstract}
Machine learning systems are often developed under data-rich conditions but
deployed when only incomplete information is available. A natural example
arises in colour fundus photography (CFP), where both eyes are commonly
imaged because they can provide complementary information for ocular disease
assessment. Prior
approaches that exploit bilateral information either fuse both eyes in a
single model or use knowledge distillation to compress a dual-eye teacher
into a lightweight student that still requires both images at test time. However, bilateral acquisition may not always be feasible. We thus propose a privileged learning framework in which a teacher
model has access to bilateral CFP images during training and learns
representations informed by information from both eyes, while the student
is trained to transfer this knowledge from the bilateral teacher using only
a single-eye image as input. At inference, the student operates exclusively
on this single-eye input and does not require the contralateral image.
Experimental results show consistent improvements in area under the
receiver operating characteristic curve over conventional monocular
training, with the largest gains observed for left-eye images for the detection of
age-related macular degeneration (0.931 vs.\ 0.966) and diabetic
retinopathy (0.789 vs.\ 0.836).%, while requiring only a single-eye image at inference. 
By enabling models to benefit from bilateral information during
training while requiring only a single-eye image at inference, our approach
may support more reliable deployment in resource-constrained clinical and
screening environments.
\end{abstract}

\textbf{\large{1 Introduction}}

Machine learning systems are often developed under data-rich conditions but deployed when only incomplete information is available. This mismatch is particularly important in clinical settings, where acquisition failures, limited infrastructure, patient-related factors, or operational constraints may prevent models from accessing all observations available during development. Consequently, models that exploit multiple complementary observations during training may face a practical challenge at deployment when one or more of these observations are unavailable or unusable.

A natural example arises in colour fundus photography (CFP), where bilateral imaging is commonly performed because the two eyes can provide complementary information for ocular disease assessment~\citep{multieye2025}. However, obtaining two usable fundus images is not guaranteed in routine screening. Fundus images can be ungradable because of media opacity, inadequate pupil dilation, defocus, poor fixation, or other acquisition-related factors, and image quality can differ between fellow eyes. Previous CFP studies have reported high but incomplete bilateral gradability; for example, approximately 92--94\% of participants had gradable images in both eyes in some population-based studies, while substantially higher rates of unilateral or bilateral ungradability have been reported in other clinical screening settings~\citep{aung2009retinal,piyasena2019mydriatic}. Thus, although bilateral imaging may be attempted and remains clinically valuable, a deployment scenario in which only one usable eye is ultimately available is realistic. Requiring bilateral images at inference would exclude such cases even when the available monocular image is itself sufficient for prediction.

This creates a distinction between models trained using monocular inputs and those trained using bilateral inputs. A monocular model can operate whenever at least one usable eye is available, but cannot directly exploit information contained in the contralateral eye. Conversely, a bilateral model can incorporate information from both eyes but requires both images to be available and usable at inference~\citep{he2020classification}. This represents a practical trade-off between access to complementary information and robustness to incomplete acquisition. In settings where repeat image acquisition, pupil dilation, or additional clinical assessment may be costly or impractical, a model that can operate from a single usable eye may provide greater deployment flexibility. This motivates the question: \textit{can information from bilateral observations available during training be transferred to a model that requires only a monocular image at inference?}

In this paper, we examine this question by leveraging a publicly available dataset and proposing a teacher-student framework in which a teacher model has access to bilateral image inputs during training and can exploit information from the contralateral eye when both images are available. A student model is subsequently trained to transfer this knowledge while receiving only monocular inputs. At inference, the student operates exclusively on a single-eye image and therefore remains applicable when the contralateral image is unavailable or ungradable. Here, we refer to information available during training but unavailable at inference as \textit{privileged information}; in our setting, the \textit{contralateral-eye image} constitutes this privileged information.

Prior  work has  explored learning from privileged
information that is available during training but unavailable at deployment.
For example, \citet{duan2014incorporating} incorporate privileged
genetic information into fundus-image-based glaucoma detection, while
\citet{korevaar2024domain} perform macular disease severity
classification using optical coherence tomography images and incorporate
fluid segmentation masks as privileged information to improve domain
generalization. Studies have also exploited bilateral fundus images for ocular disease
classification by weighting and fusing information from both
eyes~\citep{he2020classification,al2024fundus}. Such approaches can benefit
from complementary bilateral information but require paired images at
inference. More recent work has explored knowledge distillation to reduce
model size while retaining dual-eye inputs; e.g. \citet{soflaei2025explainable} propose to train a 
student whose Siamese variant fuses information from both eyes for cataract
detection, but this formulation still requires both  images at
inference. % In contrast, we investigate the contralateral eye as privileged information, with bilateral observations available during training but only a single-eye image required at inference. %Whereas distillation-based dual-eye approaches such as~\citep{soflaei2025explainable} retain the bilateral input at test time, our student is explicitly monocular at inference: the contralateral eye only shapes the representation during training and is discarded afterward.

%Prior work has exploited bilateral fundus images for ocular disease classification by weighting and fusing information from both eyes~\citep{he2020classification,al2024fundus}. Such approaches can benefit from complementary bilateral information but require paired images at inference. Related work has also explored learning from privileged information that is available during training but unavailable at deployment. For example, Dual et al.\cite{duan2014incorporating} incorporate privileged genetic information into fundus-image-based glaucoma detection, while Korevaar et al.\citep{korevaar2024domain} perform macular disease severity classification using optical coherence tomography images and incorporate fluid segmentation masks as privileged information to improve domain generalization. In contrast, we investigate the contralateral eye as privileged information, with bilateral observations available during training but only a single-eye image required at inference.

Our approach therefore seeks to combine the complementary information available in bilateral observations during development with the acquisition flexibility of monocular inference. Rather than requiring bilateral imaging at deployment, the proposed framework uses bilateral information to guide representation learning during training and transfers this information to a student model that can operate from either eye independently. This formulation provides a practical way to exploit richer training observations without imposing the same acquisition requirements on the deployed system, and shifts the benefit of dual-eye distillation from model compression
toward deployment flexibility under incomplete acquisition.

\vspace{1em}
\textbf{\large{2 Methods}}

%\subsection{Overview}
%lightweight and high-efficiency 

We build our % developed a privileged distillation 
framework using  RETFound-Green, a pretrained retinal foundation model based on the Vision Transformer  architecture~\citep{engelmann2025training}. RETFound-Green contains 22.2 million parameters and was designed as a compact, computationally efficient retinal foundation model. The word \textit{Green} in the name reflects its emphasis on reducing computational and resource requirements while maintaining competitive performance for retinal image analysis. This compact architecture makes RETFound-Green well suited to our setting, where computational efficiency and practical deployment are important considerations.

Our proposed framework consists of a teacher and a student model. The teacher model was trained on paired left and right images, learning features from both eyes and fusing them via mean pooling to form a privileged bilateral representation that was subsequently passed to a linear classifier.  A distillation loss guided the student to learn features that aligned with the teacher's richer bilateral representation. The overall training objective combined the binary cross-entropy (BCE) classification losses of both networks with a feature-level distillation loss, given by the third term:\vspace{-.85em}
\begin{equation}
    \mathcal{L} = \text{BCE}( \hat{y}_{teacher}, y ) +\text{BCE}( \hat{y}_{student}, y )  + \alpha \frac{1}{B} \sum^B_{i=1}|| \mathbf{Z}^{i}_{student} - \mathbf{Z}^{i}_{teacher} ||_2^2,
    \label{eq:total_loss}
\end{equation}
where $\mathbf{Z}^{i}_{student} \in \mathbb{R}^{384}$ and $\mathbf{Z}^{i}_{teacher} \in \mathbb{R}^{384}$ were the feature representations from the second-to-last layer of each network, extracted from the monocular images and the fused bilateral representation, respectively. The hyperparameter $\alpha$ balanced the classification and distillation objectives. We set $\alpha$ = 0.1 for all experiments in this work.  

\newpage
\textbf{2.1 Model Architectures}

We considered three model configurations.

\textbf{Baseline} ($M_{\mathrm{baseline}}$). A single-eye network trained to
classify monocular fundus images, optimized solely via the binary
cross-entropy loss. In every training batch, both eyes of each patient are
included as separate monocular examples.

\textbf{Privileged Teacher} ($M_{\mathrm{teacher}}$). A separate network
that processes paired left and right fundus images, extracts per-eye
features, mean-pools them into a bilateral representation
$\mathbf{Z}_{\mathrm{teacher}}$, and classifies this representation with a
linear head. It is trained with the classification loss only. The teacher's
parameters are not shared with the student.

\textbf{Privileged Student} ($M_{\mathrm{student}}$). Identical architecture
and training inputs to the baseline: the same single-eye network, and the
same batches containing both eyes of each patient as separate monocular
examples. The only difference from the baseline is the loss. During
training, the student's feature representation $\mathbf{Z}_{\mathrm{student}}$
is additionally matched to the teacher's detached bilateral representation
$\mathbf{Z}_{\mathrm{teacher}}$.
The teacher's representation is detached before the MSE, so
gradients from the distillation term update the student only.

Because the baseline and the privileged student share architecture, inputs,
and training schedule, the only difference between them is this additional
supervision. At inference, only the student is used, on a single eye; the
teacher is not required.

\textbf{2.2 Experimental Setup}\label{sec:setup}

\textbf{Dataset.} We developed and validated our proposed method using the Ocular Disease Intelligent Recognition (ODIR-5K) dataset~\citep{odir2019}.

\textbf{Tasks.} We considered two detection tasks: age-related macular
degeneration (AMD) and diabetic retinopathy (DR).

\textbf{Data split.} To obtain unbiased performance estimates, we partitioned the ODIR-5K dataset at the patient level, ensuring that images from the same patient never appeared in more than one data subsets. We first performed a 6-fold split to separate a held-out test set (20\% of patients). From the remaining patients, we performed a second 5-fold split to create a training set (80\%) and a validation set (20\%). This hierarchical, patient-aware splitting strategy prevented data leakage and evaluated the model's ability to generalize to new patients.

The validation set was used for (1) {early stopping} during training and (2) {threshold calibration}. At the end of each epoch, we computed the binary cross-entropy loss on the validation set and saved the model checkpoint with the lowest validation loss. For threshold calibration, we selected the optimal decision threshold \(t^*\) that maximized the macro F1-score. \iffalse 
\[
t^* = \arg\max_{t \in [0.01, 0.99]} \text{F1}_{\text{macro}}(y_{\text{val}}, \bm{1} [p_{\text{val}} > t]).
\]
where \(\bm{1}\) is the indicator function. \fi 
The threshold was then applied to the held-out test set predictions. All final metrics reported in this work were computed on the test set, which remained completely unseen during training, model selection, and threshold calibration. 

\textbf{Performance evaluation.} We computed Area Under the Receiver Operating Characteristic Curve (AUROC),  Area Under the Precision–Recall Curve (AUPRC), Balanced Accuracy, Precision, and Recall, alongside false positive rate (FPR) to transparently capture task‑specific clinical trade‑offs.

\vspace{1em}
\textbf{\large{3 Results \& Discussion}}

\begin{table*}[t!]
\centering  \small 
\setlength{\tabcolsep}{4pt}
\renewcommand{\arraystretch}{1.15}
\caption{{Performance metrics with 95\% confidence intervals for AMD (top) and DR (bottom). Bold indicates the best result for each eye and task. The privileged framework improves AUROC and balanced accuracy across both tasks, with task-specific trade-offs: increased recall for AMD (catching more positives) and increased precision for DR (reducing false positives).}}
\label{tab:core}
\begin{tabular}{lcc|cc}
\toprule \small 
\textbf{Metric} & \textbf{Baseline (Left)} & \textbf{Privileged (Left)} & \textbf{Baseline (Right)} & \textbf{Privileged (Right)} \\
\midrule
\multicolumn{5}{c}{\textbf{Task: Detection of Age-related Macular Degeneration (AMD)}} \\
\textbf{AUROC} & 0.931 (0.846--0.984) & \textbf{0.966} (0.936--0.987) & 0.801 (0.679--0.890) & \textbf{0.810} (0.710--0.889) \\
\textbf{Balanced Accuracy} & 0.834 (0.744--0.924) & \textbf{0.879} (0.799--0.957) & 0.676 (0.585--0.774) & \textbf{0.689} (0.592--0.784) \\
\textbf{AUPRC} & \textbf{0.696} (0.520--0.838) & 0.651 (0.473--0.821) & \textbf{0.446} (0.259--0.642) & 0.413 (0.221--0.610) \\
\textbf{Precision} & \textbf{0.760} (0.588--0.917) & 0.595 (0.429--0.744) & 0.\textbf{769} (0.500--1.000) & 0.579 (0.333--0.833) \\
\textbf{Recall} & 0.679 (0.500--0.857) & \textbf{0.786} (0.625--0.939) & 0.357 (0.176--0.552) & \textbf{0.393} (0.200--0.577) \\
\textbf{FPR} ($\downarrow$ is better) & \textbf{0.011} (0.004--0.020) & 0.027 (0.014--0.041) & \textbf{0.005} (0.000--0.011) & 0.014 (0.005--0.025) \\
\midrule
\multicolumn{5}{c}{\textbf{Task: Detection of Diabetic Retinopathy (DR)}} \\
\textbf{AUROC} & 0.789 (0.734--0.843) & \textbf{0.836} (0.799--0.872) & 0.825 (0.787--0.860) & \textbf{0.837} (0.801--0.871) \\
\textbf{Balanced Accuracy} & 0.576 (0.548--0.601) & \textbf{0.749} (0.712--0.788) & 0.740 (0.703--0.776) & \textbf{0.757} (0.719--0.791) \\
\textbf{AUPRC} & 0.511 (0.417--0.608) & \textbf{0.772} (0.717--0.824) & 0.759 (0.705--0.811) & \textbf{0.778} (0.724--0.827) \\
\textbf{Precision} & 0.078 (0.064--0.094) & \textbf{0.760} (0.688--0.831) & 0.741 (0.669--0.807) & \textbf{0.803} (0.733--0.867) \\
\textbf{Recall} & \textbf{0.917} (0.865--0.966) & 0.587 (0.516--0.659) & 0.577 (0.508--0.643) & \textbf{0.582} (0.511--0.650) \\
\textbf{FPR} ($\downarrow$ is better) & 0.766 (0.744--0.787) & \textbf{0.089} (0.061--0.117) & 0.096 (0.069--0.127) & \textbf{0.068} (0.044--0.094) \\
\bottomrule
\end{tabular}
\end{table*}

\textbf{Disease prevalence}. The developmental dataset contains 136 positive cases for AMD and 939 positive cases for DR, corresponding to a prevalence of 4.8\% and 32.4\%, respectively. The held-out test set contains 28 positive AMD cases (4.7\%) for AMD and 189 positive cases for DR cases (32.2\%). 
 
\textbf{Performance \& trade-offs.} Table \ref{tab:core} summarizes the results across all experimental trials. We report 95\% confidence intervals for all metrics, estimated via bootstrapping on the held-out test set to quantify sampling variability. The privileged-distillation framework improved AUROC and balanced accuracy for the left eye (0.931 vs. 0.966 and 0.834 vs. 0.879, respectively), indicating improved ranking performance and class-balanced accuracy. However, this improvement was accompanied by a higher false-positive rate (0.011 vs. 0.027) and lower precision (0.760 vs. 0.595).

This trade-off reflects a shift toward higher sensitivity (recall: 0.679 vs. 0.786), which may be desirable in screening applications where false negatives can be more costly than false positives \cite{wang2024economic}. The increase in FPR remains modest in absolute terms (from 1.1\% to 2.7\%), while the improvement in balanced accuracy indicates improved aggregate performance across the two classes. Nevertheless, the accompanying reduction in precision highlights a potential cost of this increased sensitivity.

For DR, the privileged model similarly improved AUROC and balanced accuracy across both eyes, with the largest improvement observed for the left eye (AUROC: 0.789 vs. 0.836; balanced accuracy: 0.576 vs. 0.749). %, although threshold-dependent metrics exhibited a different precision--recall trade-off.
Unlike AMD, the left-eye DR model achieved substantially higher precision and lower false-positive rate at the expense of recall.
 
\textbf{Limitations.} First, the relatively small test set limits the statistical power and precision of our estimates and therefore the strength of our generalizability claims. External validation is needed to assess robustness across diverse populations, acquisition conditions, and imaging systems. Second, while the proposed method improves classification performance, clinical deployment would require careful integration with existing screening workflows, including appropriate human oversight and evaluation of model interpretability. Finally, our experiments were conducted using a single dataset split. Evaluation across multiple random seeds and, where feasible, cross-validation would provide a more robust assessment of performance variability.

\vspace{1em}
\textbf{\large {4 Conclusions}}

We introduced a privileged distillation framework for retinopathy detection using monocular images, where a teacher network trained on paired fundus images guides a student network to process only monocular images. The proposed method was  evaluated against a baseline that treats each eye independently. %Using a patient‑wise split and a comprehensive set of metrics with bootstrapped confidence intervals.
Our framework offers a practical, deployable solution for DR screening, particularly in settings where only one fundus image is available. The observed trade‑offs and asymmetries reveal insights for future research into multimodal and asymmetric medical imaging.

\appendix
\newpage
\section{Feature Quality via Linear Probing.}\label{sec:ablation}
  
To determine whether the teacher improved the intrinsic quality of the student's features (beyond the final classification head), we performed linear probing \citep{alain2016understanding}. We froze the backbones of both the baseline and privileged student models and replaced their classification heads with a fresh, randomly initialized linear layer. We then trained \textit{only} this new linear layer on the training set for 20 epochs, while keeping the backbones entirely frozen. The performance (AUROC) of this probe was evaluated on the test set. 
If the privileged backbone achieves a higher AUROC than the baseline backbone under this protocol, it provides direct evidence that the teacher improved the linear separability and quality of the student's learned features.

\subsection{Results} 

Linear probing revealed that the privileged student's frozen backbone achieved a +2.0 percentage-point AUROC improvement over the baseline backbone for AMD when trained with a new linear classifier (0.815 vs. 0.836). A similar +2.9 percentage-point improvement was observed for DR. These results indicate that disease-relevant information was more linearly accessible from the privileged student's frozen representations than from the baseline representations. This suggests that the benefits of privileged-information distillation extend beyond optimization of the original classification head and are reflected in the learned backbone representations.

{
\small
\bibliography{iclr2025_conference}
\bibliographystyle{iclr2025_conference}
}

\end{document}